\documentclass[10pt,twocolumn,letterpaper]{article}

\usepackage{cvpr}
\usepackage{times}
\usepackage{epsfig}
\usepackage{graphicx}
\usepackage{amsmath,amssymb,amsfonts}
\usepackage{algorithmic}
\usepackage{subcaption}
\usepackage{textcomp}
\usepackage{xcolor}
\usepackage{pgfplots}
\pgfplotsset{width=0.9\columnwidth, height=5cm}

\usepackage[breaklinks=true,bookmarks=false]{hyperref}

\cvprfinalcopy 

\def\cvprPaperID{****} 

\begin{document}

\title{Understanding Deep Convolutional Networks through Gestalt Theory}

\author{Angelos Amanatiadis\\
Democritus University of Thrace\\
{\tt\small aamanat@ee.duth.gr}
\and
Vasileios G. Kaburlasos\\
EMaTTech\\
{\tt\small vgkabs@teiemt.gr}
\and
Elias B. Kosmatopoulos\\
Democritus University of Thrace\\
{\tt\small kosmatop@ee.duth.gr}
}

\maketitle

\begin{abstract}
   The superior performance of deep convolutional networks over high-dimensional problems have made them very popular for several applications. Despite their wide adoption, their underlying mechanisms still remain unclear with their improvement procedures still relying mainly on a trial and error process. We introduce a novel sensitivity analysis based on the Gestalt theory for giving insights into the classifier function and intermediate layers. Since Gestalt psychology stipulates that perception can be a product of complex interactions among several elements, we perform an ablation study based on this concept to discover which principles and image context significantly contribute in the network classification. Our results reveal that convnets follow most of the visual cortical perceptual mechanisms defined by the Gestalt principles at several levels. The proposed framework stimulates specific feature maps in classification problems and reveal important network attributes that can produce more explainable network models.
\end{abstract}

\section{Introduction}

Explaining how deep neural networks work is an open and challenging task. Certain insights about their underlying mechanisms can help users and researches to explain their operation, identify potential strengths and weaknesses, and help them to easily understand how they work, operate and infer. In addition, network optimization will be an easier process for the user, as it will be easier to identify errors and imperfections in terms of architecture and accuracy.\par

\begin{figure}
	\begin{center}
	\includegraphics[width=0.48\textwidth]{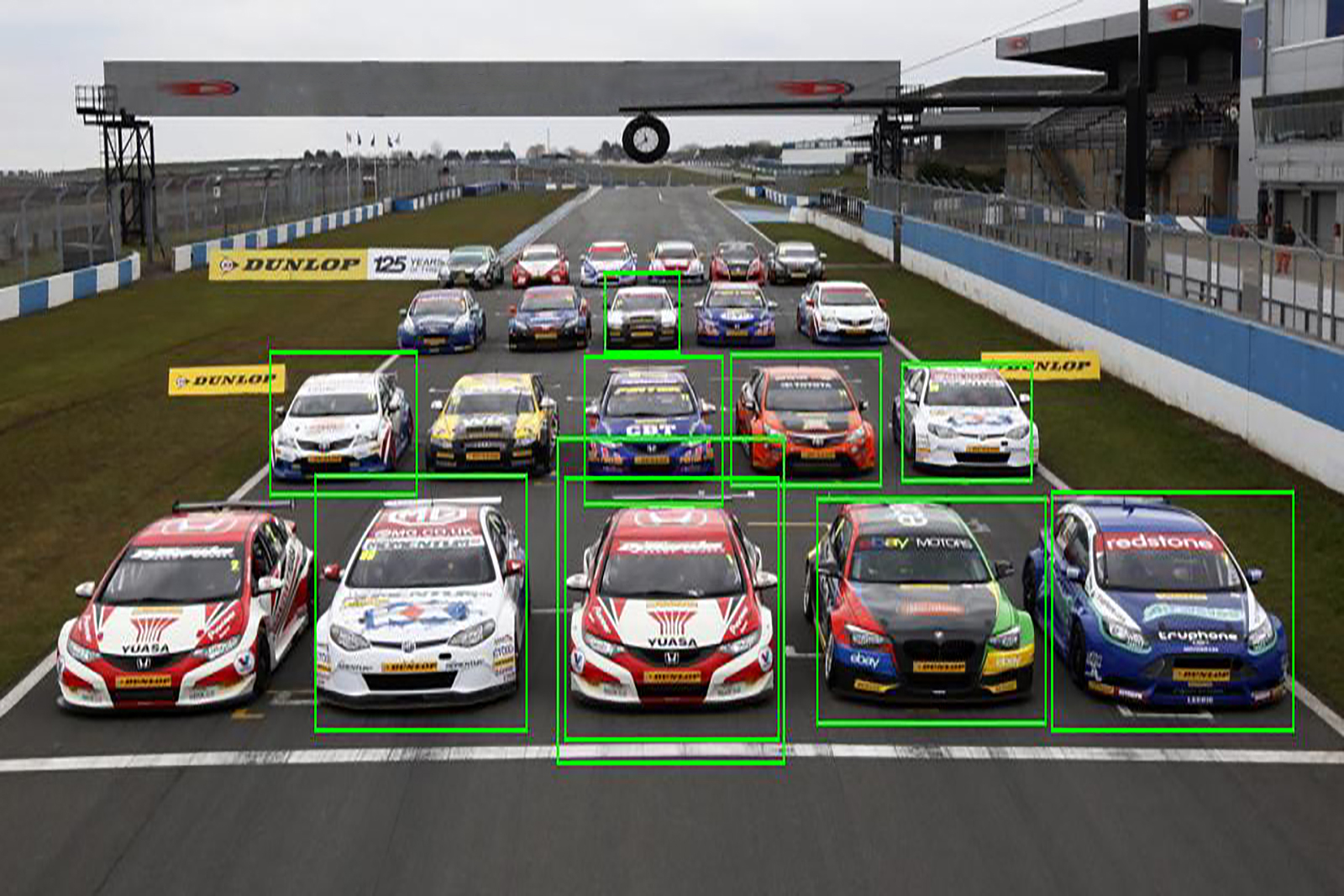}
	\end{center}
\caption{A simplified example of vehicle detection results using a convolutional neural network. While the results are partially correct, at the same time many questions also arise to the user about the network's operation.}
\label{cnnexample}
\end{figure}
The need for better model understanding has become even more urgent, as the use of deep learning techniques is already extended to areas such as autonomous vehicles, robotic industrial systems, medical devices, and security systems. Inevitably, in all these systems there will be a human interaction (autonomous vehicle - pedestrian, industrial robot - worker) resulting in critical decisions for the autonomous systems which can even influence the lives of people interacting with them. These questions and concerns have been established and have recently become widely known following various open letters co-signed by distinguished scientists and academics \cite{russell2015letter}. The better understanding of the emerging generation models of deep convolutional networks will help the scientific community to bring further developments with the ultimate goal in the end-user confidence.\par
Various works have been recently proposed for such understanding targeting at different methodologies and goals. A visualization method by projecting  the feature activations back to the input pixel space has been proposed, showing which patterns from the training set activate the feature map \cite{zeiler2014visualizing}. This approach utilizes a fully supervised convolutional network model together with deconvolutional network for the necessary mapping. A framework for explaining the transfer learning of deep networks by analyzing their contraction and separation properties have been recently also proposed \cite{mallat2016understanding}. This explanation was achieved by separating variations of certain operators with a wavelet transform at different scales. Insights about the captured invariances by deep network representations are also given in \cite{mahendran2015understanding} where a progressive and more invariant and abstract image notions are formed throughout the network \cite{charalampous2012sparse}. Using invertible deep networks is another approach for such analysis since several properties can be back-tracked from the feature space to the input space \cite{jacobsen2018revnet}.\par
The purpose of this research is to help the end user understand in a more practical way the rationale behind the decisions made by the used deep convolutional network. Decisions, recommendations, classifications, or actions produced by such a system should be as comprehensible as possible to the human user so that he can more easily trust and evaluate them. This is achieved by various sensitivity analyses based on Gestalt principles aiming to define how deep networks perceive patterns and organize the raw input data.

\section{Motivation and Gestalt Theory}
\begin{figure}
	\begin{center}
	\includegraphics[width=0.48\textwidth]{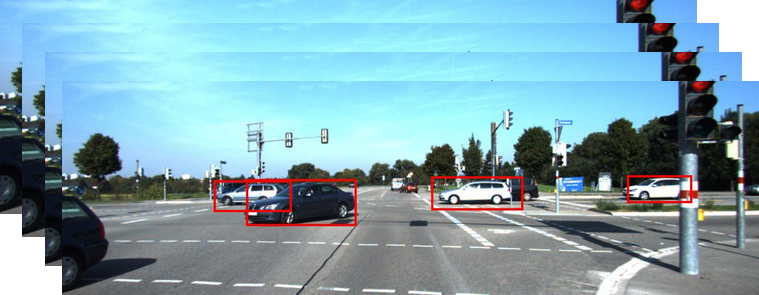}
	\end{center}
\caption{Training images for vehicle detection along with their ground truth bounding boxes.}
\label{kitty}
\end{figure}

In Figure \ref{cnnexample}, an example of vehicle detection results is depicted. The system is required to detect all vehicles which are included in a single image. A typical Convolutional Neural Network (CNN) architecture was used, pre-trained using the ImageNet \cite{deng2009imagenet} image database. Afterwards, the network is trained on the KITTI dataset \cite{geiger2013vision}, which contains $7481$ car pictures in various colors, scales, and orientations, as in Fig. \ref{kitty}. After the training, an image query is introduced to the detection network, with its results presented in Fig. \ref {cnnexample}, where successful car detection is represented by a green bounding box surrounding the detected vehicle.\par
By analyzing the results, the user can conclude that the trained network is partially successful in detecting vehicles, but at the same time, many questions about its internal functionality emerge. Some questions that might arise by the user can include: Why the system did not detect all the cars? Why the left vehicle of the first row was not detected while several cars in other back rows were successfully detected?  Does the color of a car contribute to a successful detection? If so, why the front row left car was not detected while the same car with the same colors in the middle of the row was successfully detected? Does adding convolutional layers make the network more efficient in detecting more cars? Will I get better detection results if I use more training image samples than the initial $7481$ ones?\par

Many of the aforementioned questions have been discussed in several works with concrete answers mostly in the feature vector and visualization domains. A widely adopted solution is to perform several trial and errors in the network architecture and the training dataset to better answer the detection questions. However, even in such time-consuming trials, some insights of the network's mechanism are still elusive.\par
\begin{figure}
	\begin{center}
	\includegraphics[width=0.48\textwidth]{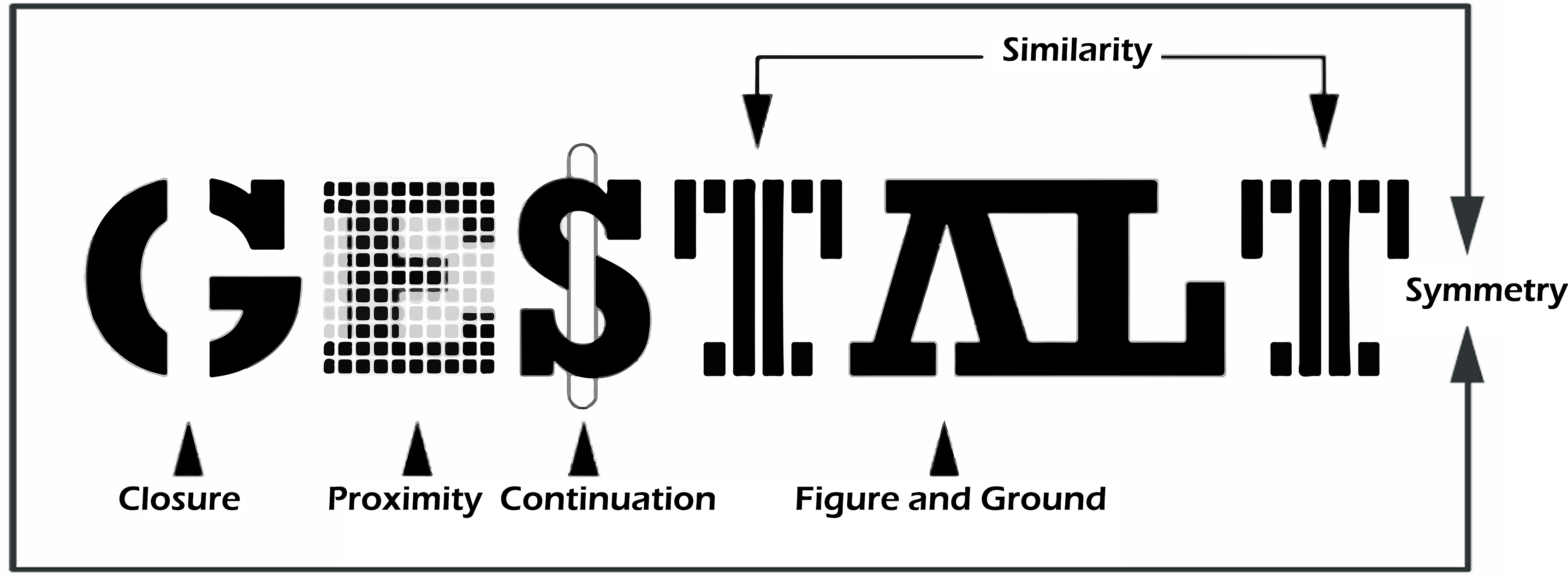}
	\end{center}
\caption{Gestalt principles that can be used to explain how deep convolutional networks perceive certain patterns. (Image source: Wikipedia)}
\label{intro}
\end{figure}

It is widely known that CNNs is a biologically inspired technique trying to resemble the visual cortex of the human brain. By definition there is no clear analogy between a CNN and the visual cortex functionally, however, CNNs try to replicate the efficiency and robustness by which the human brain represents visual information. In this respect, Gestalt Psychology tries also to understand how humans understand and see forms. This theory can be summed up by the single quote that ``\textit{the whole is different from the sum of its parts}'', meaning that human brain identifies the totality of something before grasping all the details \cite{koffka2013principles}. In other words, the human perceptual system perceives a whole independently of its parts. The Gestalt theory developed some principles to explain how patterns are perceived and how the human brain handles the sensed raw data from our senses \cite{wertheimer1923untersuchungen}. Various studies based on the Gestalt theory have been also presented in the domain of image understanding \cite{bileschi2007image}, \cite{zlatoff2004image}. A brief discussion of these principles, which are summarized in Fig. \ref{intro}, and their potential correlation to CNN behavior is provided.

\subsection{Closure}  
The principle of closure supports that humans can perceive several objects even in the case of an incomplete structure. This visual gap is filled by the human perception and is triggered when picture parts are missing. The missing information is filled by lines, color or patterns so as to complete the image. For example, in Fig. \ref{intro}, the letter ``G'' is not complete but with two separate gaps. If this principle was not applied, the image would be perceived as an assortment of three lines with different lengths, rotations, and curvatures, however, the lines are perceptually combined to form the letter ``G''.

\subsection{Proximity}
The proximity principle declares that elements with spatial closeness are grouped together forming a single entity. In Fig. \ref{intro}, the letter ``E'' is illustrated with 74 small squares but we perceive this collection of squares in a group. This perception suggests that the brain builds connections between similar elements. Similarly, elements that are far from others can form a different part of an object with the defining distance being subjective. This grouping of smaller sets into larger ones unburdens the human brain from processing too many and small stimuli.
 
\subsection{Continuation} 
The next principle, continuation, is the concept of even though we don't actually see shapes continuing, it is strongly implied. In other words, the mind can continue a pattern even after the pattern stops. In the example Fig. \ref{intro}, the letter ``S'' includes a long line with missing parts. Actually, this is not a single line, but three separate ones. The principle of continuity is applied not only for single lines but any pattern can be used to be integrated as a perceptual whole if there are some rules of alignment. Even in object intersection, the human brain tends to perceive these two objects as two single uninterrupted entities. Elements with sharp abrupt directional changes are less likely to be grouped as a single object. 
 
\subsection{Similarity} 
Based on the similarity principle, objects that are similar with certain attributes are more likely to be grouped together. This similarity relies on various relationships such as form, color, size, and brightness. For example, objects of the same color or size are more possible to be seen as a group. Consequently, different objects can be made look similar, if something common like color or shape is introduced to them. These relationships are built in the human brain and crafts links between similar nature's element for separating them from other elements.

\subsection{Figure and Ground}
The Figure and Ground principle implies that certain objects in a figure have a more prominent role and are first perceived. In the example Fig. \ref{intro}, a tree lies inside the letter ``A'', however based on the focus and the background the letter ``A'' is first perceived. In a visual scene, the brain decides which object is the figure in focus and which other figures are part of the background. This discriminative decision depends on many cues, following a probabilistic nature. Color, shape, edges, and size inherently contribute to this probabilistic process, with the brain considering all these cues for a final softmax decision.   
\subsection{Symmetry}
The symmetry principle states that when symmetries are apparent in a figure, the mind tends to group certain objects based on these symmetries. This implies that the mind tries to form objects around a center point and locate symmetries for creating perceptual connections of a coherent shape. When similarities exist between symmetrical objects, the grouping likelihood is further increased regardless of their distance. Finally, symmetry is considered stronger that proximity.


\section{Approach}
Based on the Gestalt principles we define the approach for our attempt to divulge some of the properties of CNNs. For the closure principle, we will train a CNN on a digit or letter dataset and query several digits or letters with missing information, respectively. For evaluating potential proximity properties several letters or digits can be formed as small elements with spatial closeness and introduced to the networks as queries. The proximity distance can be also evaluated by considering and modifying the spatial closeness of the elements. For the continuation rule, several items and patterns with and without abrupt directional changes will be evaluated for possible detection in a more deep network architecture trained on the ImageNet database. An approach to evaluate the similarity in a CNN, is to try to fool a detection network with different objects but with similar  attributes for grouping them together. For the figure and ground principle, a convnet trained on the ImageNet will be used to evaluate how a figure is perceived based on the key aspects of color, shape, and edges. For this evaluation, specific images that exemplify these key aspects will be used to try to understand the CNN interpretations. Finally, symmetry will be assessed utilizing pairs of shapes in different orientations until a symmetry can be achieved leading to a potential successful detection.\par
For evaluating the principles certain types and architectures of convolutional networks have to be chosen.  We chose standard and widely used fully supervised convnet models with no transfer learning. More precisely, for the closure and proximity, the chosen network (AlexNet) is described in \cite{krizhevsky2012imagenet} trained in the MNIST handwritten digit database \cite{lecun1998mnist}. For the rest of the principles, the Inception V1 (GoogLeNet) \cite{szegedy2015going} was utilized trained on the ImageNet \cite{deng2009imagenet} image database. A softmax classifier is used by both networks evaluating the feature representations learned from MNIST and ImageNet datasets, respectively. Each principle will be evaluated on specific image attributes with several sensitivity analyses based on parameters that are correlated to the Gestalt theory. \par

\begin{figure}
\centering
\begin{subfigure}{.31\linewidth}
  \centering
  \includegraphics[width=1.0\linewidth]{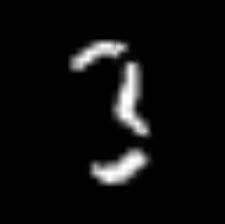}
  \caption{}
  \label{fig:sub1}
\end{subfigure}
\begin{subfigure}{.31\linewidth}
  \centering
  \includegraphics[width=1.0\linewidth]{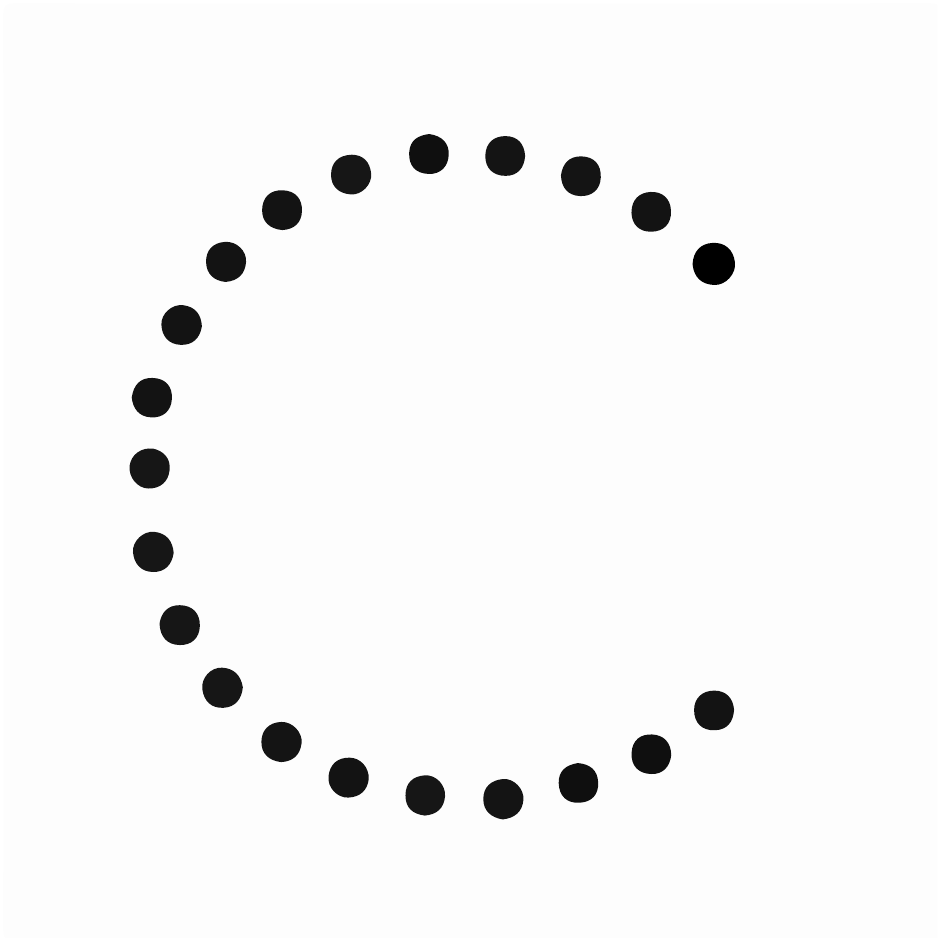}
  \caption{}
  \label{fig:sub2}
\end{subfigure}
\begin{subfigure}{.31\linewidth}
  \centering
  \includegraphics[width=1.0\linewidth]{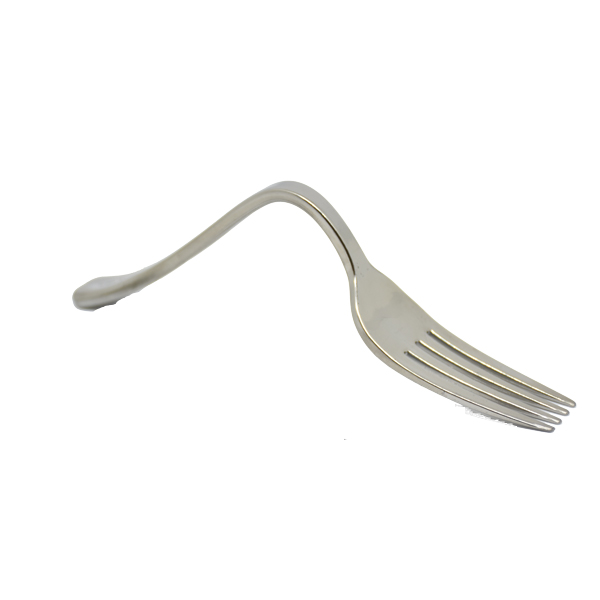}
  \caption{}
  \label{fig:sub2}
\end{subfigure}
\\[2ex]
\begin{subfigure}{.31\linewidth}
  \centering
  \includegraphics[width=1.0\linewidth]{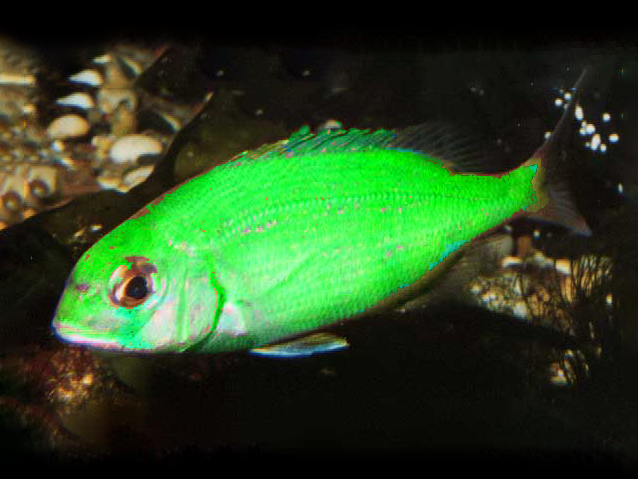}
  \caption{}
  \label{fig:sub1}
\end{subfigure}
\begin{subfigure}{.31\linewidth}
  \centering
  \includegraphics[width=1.0\linewidth]{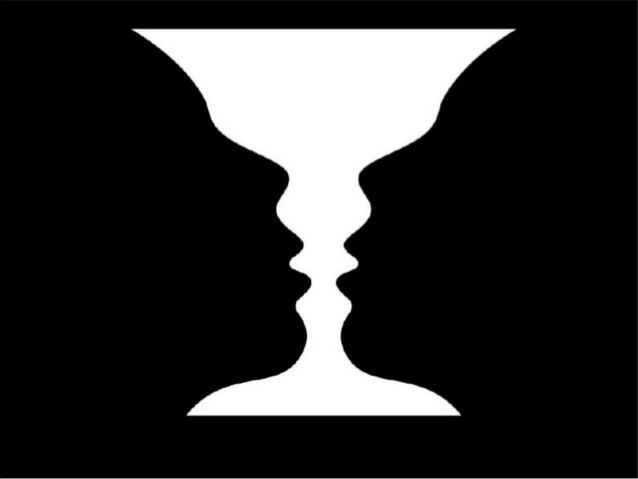}
  \caption{}
  \label{fig:sub2}
\end{subfigure}
\begin{subfigure}{.31\linewidth}
  \centering
  \includegraphics[width=1.0\linewidth]{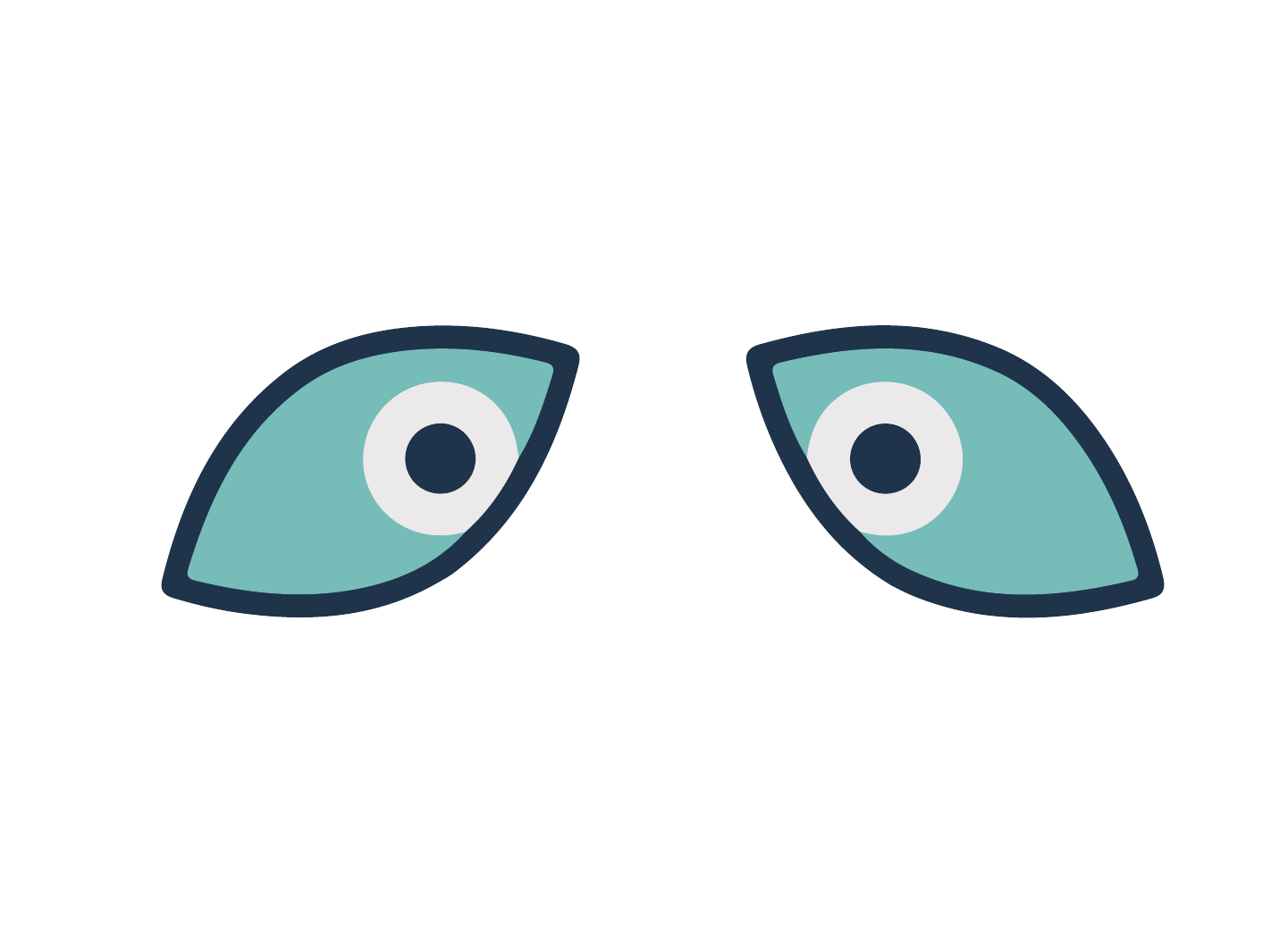}
  \caption{}
  \label{fig:sub2}
\end{subfigure}
\caption{Sample query images for evaluating whether Gestalt principles are active in convnet features. a) Closure; b) Proximity; c) Continuation; d) Similarity; e) Figure and Ground; f) Symmetry.}
\label{fig:test}
\end{figure}

\begin{figure}
	\begin{center}
	\includegraphics[width=0.48\textwidth]{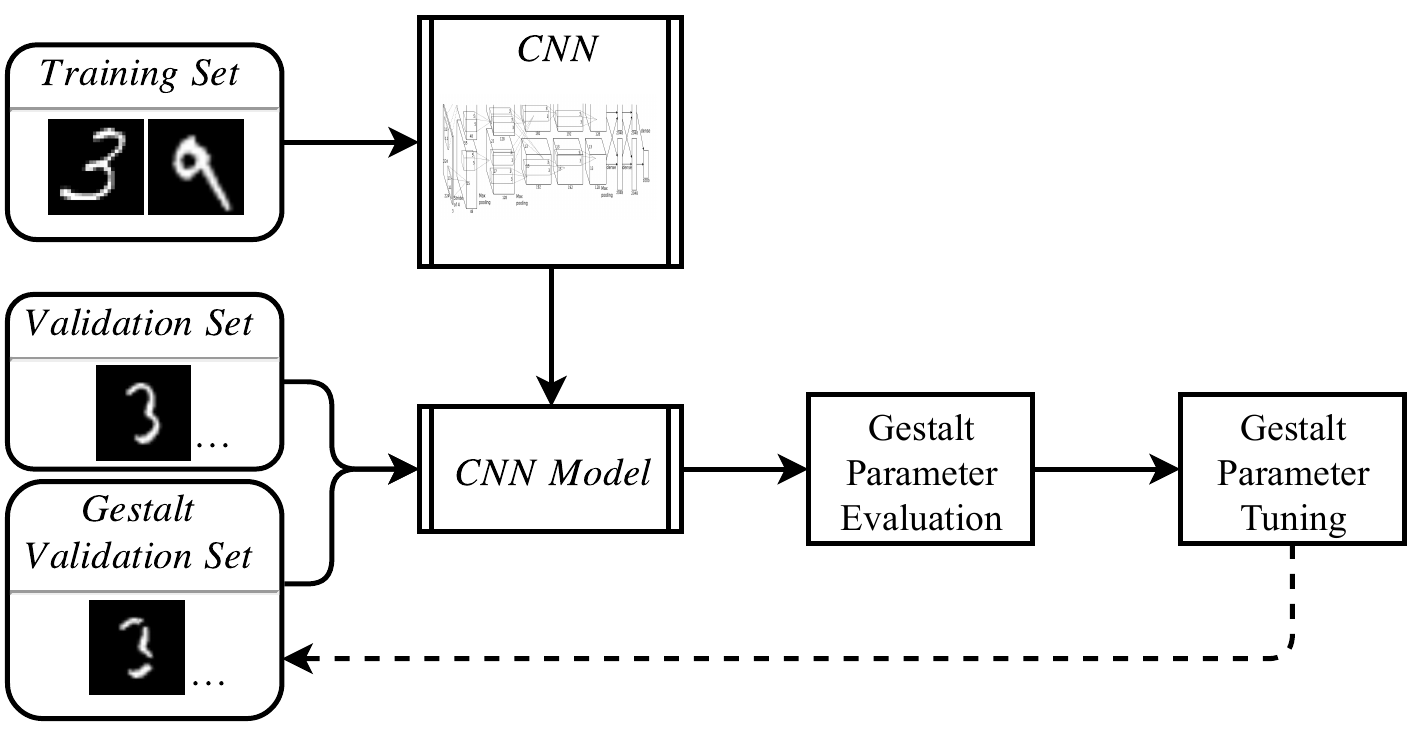}
	\end{center}
\caption{Experimental framework of the proposed evaluation method. For each Gestalt principle and parameter, different validation sets are generated based on the initial training set.}
\label{diagram}
\end{figure}

For both the convolutional neural networks we define a mapping function $f:\mathcal{R}^{N}\rightarrow\mathcal{R}^{C}$ for mapping an image $I$ to a confidence vector $y={[y_{1}, y_{2}, \dots, y_{C}]}^T$, where $C$ denotes the number of classes and $y_{i}$ the classification score of $I$ for the $i-th$ class. Unlike conventional classification works, a second validation set is used to explore the Gestalt properties on the evaluated CNN. This validation set includes images or image regions which encapsulate one specific Gestalt principle based on a single parameter for each evaluation. This gestalt parameter defined as $g$ allows us to examine the correlation of each principle in the evaluated CNN. This is achieved by comparing the accuracy of the original validation set to the gestalt validation set. Given a validation set $\mathcal{V}=\{{I_{k}\}^{M}_{k=1}}$ the accumulated accuracy of the network is defined as $h(f(\mathcal{V}))$ and the accuracy of the Gestalt validation set as $h(f(\mathcal{V},g))$ based on the calculated confidence vector. Since the Gestalt parameter is evaluated for different values during the sensitivity analysis, a Gestalt parameter detection function can be defined as:

\begin{equation} 
\label{gparameter}
g^{*}=\max_{r}(\Phi_{f}(\mathcal{V},g))
\end{equation}
where
\begin{equation} 
\label{gparameter}
\Phi_{f}(\mathcal{V},g)=h(f(\mathcal{V}))-h(f(\mathcal{V},g))
\end{equation}
where the $\Phi_{f}(\mathcal{V},g)$ compares both validation sets. With the proposed approach, as shown in Fig. \ref{diagram}, we can detect possible large accuracy drops caused by the evaluated Gestalt parameter $g$. Depending on the Gestalt principle evaluated, several rational intuitions about the CNN's behavior can be shared.

Each Gestalt parameter will define if the equivalent principle is present in the CNN and the $g^{*}$ from Eq. \ref{gparameter}, declares the maximum value for which the principle is not violated. For evaluating the closure principle, a simple patch occlusion method could be elaborated. However, since we want to measure the exact occlusion percentage on the digit shape, a fitting model was utilized for more accurate percentage occlusion. More precisely, digit shapes were represented using a free form deformation grid \cite{taron2005modelling} allowing a scale-space decomposition for accurate closure percentage estimation. For the proximity principle, the handwritten digits were firstly thinned and converted to a skeleton image composed of junction points and segments. All segments are then converted to dots whose diameter is fixed based on the corresponding area on the image grid but with a varying distance. For non-straight segments, the distance was measured along the curve \cite{gotsman1996metric}.\par
The rest four principles are evaluated against a gestalt validation set derived from the ImageNet \cite{deng2009imagenet} image database. More precisely, for the continuation evaluation, images were transformed by applying a piecewise affine transformation at a particular location. Since we want to have a continuous object partially transformed, the transformation was not applied in the whole image but in the half one. The displacement vector for the affine transformation can be defined as:

\begin{equation} 
\label{affine1}
u(x,y,\Delta \textbf{a}) = \Delta a_{0} + \Delta a_{1}x + \Delta a_{2}y
\end{equation}
\begin{equation} 
\label{affine2}
\upsilon(x,y,\Delta \textbf{a}) = \Delta a_{3} + \Delta a_{4}x + \Delta a_{5}y
\end{equation}
where $\Delta \textbf{a}=(\Delta a_{0}, \Delta a_{1}, \ldots, \Delta a_{5})$ is our gestalt parameter for this principle, measured by the affine distance \cite{werman1995similarity}. The similarity principle relies on various parameters such as form, color, size, and brightness. We evaluated it based on the color parameter by replacing the object's color in the validation set. For achieving this, a semantic segmentation if firstly applied \cite{chen2018deeplab} and then a color replacement based on the color wheel \cite{bays2009precision} is applied to the segmented object for trying to fool the CNN \cite{nguyen2015deep}. The parameter is defined as the radial angle of the selected color from the wheel with $180^{0}$ denoting an opposite color. For the figure and ground, the Gestalt validation set includes fully segmented images along with the number of classes. During the sensitivity analysis, the number of classes is reduced until the correct class is identified. Finally, for the symmetry, discrete local symmetries are firstly identified \cite{masuda1993detection}. At both symmetric points of the local symmetry, two opposite rotations are applied by an angle $\theta$ represented by:
\begin{equation} 
\label{rotation}
f(\textbf{u})=f(R(\theta)\textbf{x})
\end{equation}
where 
\begin{equation} 
\label{rotation2}
R(\theta)=
\begin{bmatrix}
    cos\theta & sin \theta \\
    -sin \theta & cos \theta
  \end{bmatrix}
\end{equation}
Since symmetries do not always exist in images found in ImageNet, this Gestalt validation set included a limited number of images. The summary of the defined Gestalt parameters for each principle is listed in Table \ref{tab1}.

\begin{table}[]
\caption{Gestalt Parameter for Each Evaluated Principle}
\begin{center}
\def\arraystretch{1.5}
\begin{tabular}{ll}
\hline \hline
\textbf{Principle} & \textbf{Gestalt Parameter}      \\ \hline \hline
Closure            & Occlusion Percentage            \\ \hline
Proximity          & Point Distance                  \\ \hline
Continuation       & Piecewise Transformation Vector \\ \hline
Similarity         & Radial Angle of Color Wheel     \\ \hline
Figure and Ground  & Number of Segmented Classes     \\ \hline
Symmetry           & Rotation Angle                  \\ \hline \hline
\end{tabular}
\end{center}
\label{tab1}
\end{table}



\section{Experimental Results}
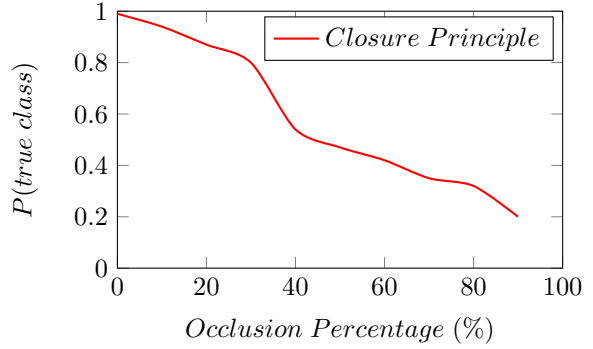
\begin{figure}
\centering
\begin{tikzpicture}
\begin{axis}[
    axis lines = box,
    xlabel = $Occlusion\;Percentage\;(\%)$,
    ylabel = {$P(true\;class)$},
		xmin=0, xmax=100,
    ymin=0, ymax=1,
		ytick={0,0.2,0.4,0.6,0.8,1},
]
\addplot [smooth,
    domain=-10:10, 
    samples=100, 
    color=red,
    	thick,
]
 coordinates {
    (0,0.99)(10,0.94)(20,0.87)(30,0.80)(40,0.54)(50,0.47)(60,0.42)(70,0.35)(80,0.32)(90,0.2)
    };
\addlegendentry{$Closure\;Principle$}
\end{axis}
\end{tikzpicture}
\caption{Probability of true label in the Gestalt validation set for closure principle.} \label{plot1}
\end{figure}

In our experiments, we first explored the closure principle. The results are shown in Fig. \ref{plot1}, where they suggest that the closure principle is found in the CNN but for an approximate threshold of about $30\%$ occlusion. After this value, a sudden drop is noticed indicating that above $30\%$ of occlusion the model cannot retain the learned local structures of the image context. In the proximity principle, as shown in Fig. \ref{plot2}, the convnet model generalizes well for dotted digits with proximity less than $5$ mm. We can understand from the results that the proximity principle is also enabled in the convets but is violated after a threshold.\par

\begin{figure}
\centering
\begin{tikzpicture}
\begin{axis}[
    axis lines = box,
    xlabel = $Point\;Distance\;(mm)$,
    ylabel = {$P(true\;class)$},
		xmin=0, xmax=20,
    ymin=0, ymax=1,
		ytick={0,0.2,0.4,0.6,0.8,1},
]
\addplot [smooth,
    domain=-10:10, 
    samples=100, 
    color=blue,
    	thick,
]
 coordinates {
    (0,0.99)(2,0.96)(4,0.92)(6,0.70)(8,0.40)(10,0.20)(12,0.15)(14,0.11)(16,0.10)(18,0.08)(20,0.06)
    };
\addlegendentry{$Proximity\;Principle$}
\end{axis}
\end{tikzpicture}
\caption{Probability of true label in the Gestalt validation set for proximity principle.} \label{plot2}
\end{figure}
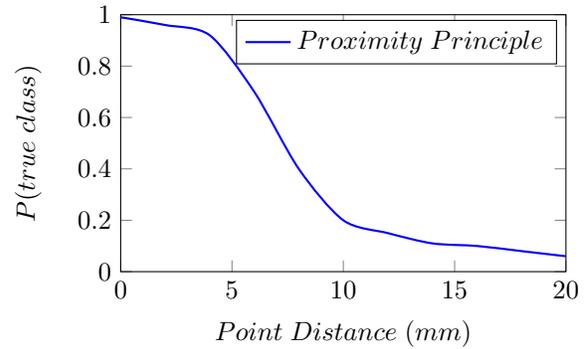

\begin{figure}
\centering
\begin{tikzpicture}
\begin{axis}[
    axis lines = box,
    xlabel = $Affine\;Distance$,
    ylabel = {$P(true\;class)$},
		xmin=0, xmax=1,
    ymin=0, ymax=1,
		ytick={0,0.2,0.4,0.6,0.8,1},
]
\addplot [smooth,
    domain=-10:10, 
    samples=100, 
    color=green,
    	thick,
]
 coordinates {
    (0,0.60)(0.1,0.58)(0.2,0.54)(0.30,0.53)(0.40,0.54)(0.50,0.57)(0.60,0.48)(0.70,0.55)(0.80,0.42)(0.90,0.39)
    };
\addlegendentry{$Continuation\;Principle$}
\end{axis}
\end{tikzpicture}
\caption{Probability of true label in the Gestalt validation set for continuation principle.} \label{plot3}
\end{figure}
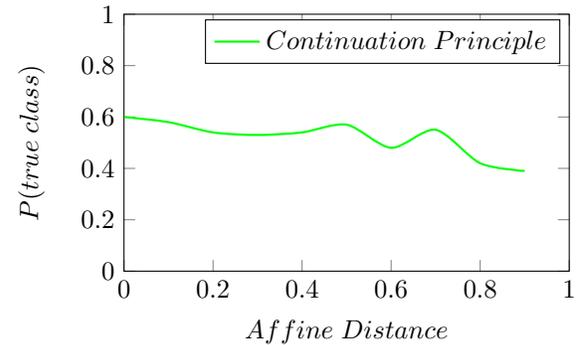

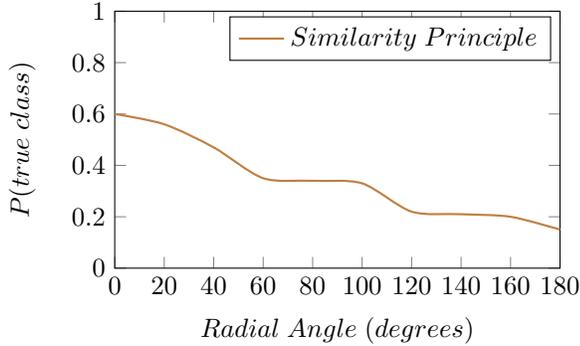
\begin{figure}
\centering
\begin{tikzpicture}
\begin{axis}[
    axis lines = box,
    xlabel = $Radial\;Angle\;(degrees)$,
    ylabel = {$P(true\;class)$},
		xmin=0, xmax=180,
    ymin=0, ymax=1,
		ytick={0,0.2,0.4,0.6,0.8,1},
		xtick={0,20,40,60,80,100,120,140,160,180},
]
\addplot [smooth,
    domain=-10:10, 
    samples=100, 
    color=brown,
    	thick,
]
 coordinates {
    (0,0.6)(20,0.56)(40,0.47)(60,0.35)(80,0.34)(100,0.33)(120,0.22)(140,0.21)(160,0.20)(180,0.15)
    };
\addlegendentry{$Similarity\;Principle$}
\end{axis}
\end{tikzpicture}
\caption{Probability of true label in the Gestalt validation set for similarity principle.} \label{plot4}
\end{figure}

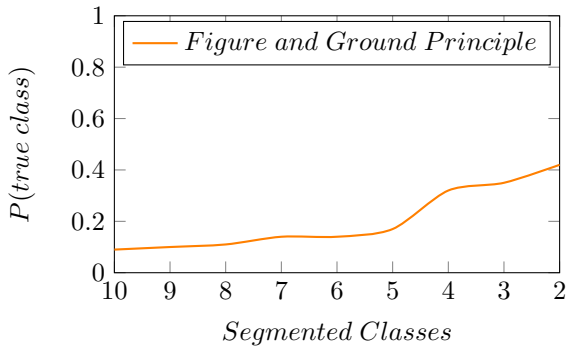
\begin{figure}
\centering
\begin{tikzpicture}
\begin{axis}[
    axis lines = box,
    xlabel = $Segmented\;Classes$,
    ylabel = {$P(true\;class)$},
		xmin=2, xmax=10,
    ymin=0, ymax=1,
		ytick={0,0.2,0.4,0.6,0.8,1},
		xtick={10,9,8,7,6,5,4,3,2},
		x dir=reverse
]
\addplot [smooth,
    domain=10:2, 
    samples=100, 
    color=orange,
    	thick,
]
 coordinates {
    (10,0.09)(9,0.10)(8,0.11)(7,0.14)(6,0.14)(5,0.17)(4,0.32)(3,0.35)(2,0.42)
    };
\addlegendentry{$Figure\;and\;Ground\;Principle$}
\end{axis}
\end{tikzpicture}
\caption{Probability of true label in the Gestalt validation set for figure and ground principle.} \label{plot5}
\end{figure}

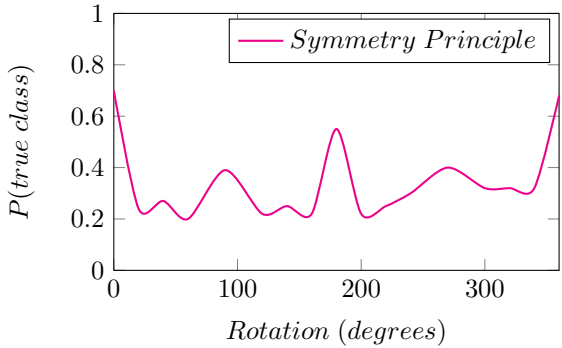
\begin{figure}
\centering
\begin{tikzpicture}
\begin{axis}[
    axis lines = box,
    xlabel = $Rotation\;(degrees)$,
    ylabel = {$P(true\;class)$},
		xmin=0, xmax=360,
    ymin=0, ymax=1,
		ytick={0,0.2,0.4,0.6,0.8,1},
]
\addplot [smooth,
    domain=-10:10, 
    samples=100, 
    color=magenta,
    	thick,
]
 coordinates {
    (0,0.70)(20,0.24)(40,0.27)(60,0.20)(90,0.39)(120,0.22)(140,0.25)(160,0.22)(180,0.55)(200,0.22)(220,0.25)(240,0.30)(270,0.40)(300,0.32)(320,0.32)(340,0.32)(360,0.68)};
\addlegendentry{$Symmetry\;Principle$}
\end{axis}
\end{tikzpicture}
\caption{Probability of true label in the Gestalt validation set for symmetry principle.} \label{plot6}
\end{figure}

In the results of the continuation principle, as shown in Fig. \ref{plot3}, the network output seems to be quite stable with a small drop in the performance. The are no dramatic effects in the performance leading to a suggestion that the continuation could be not applicable in the convnets. A possible explanation to this stability might be the fact that only half of the image was transformed, and combined with the results of closure, the continuation principle is not very strong in convnet inference. The similarity principle, as discussed previously, was evaluated against the color. From the low drop in the performance of Fig. \ref{plot4}, we can understand that color information is processed by the convnet. However, deeper networks might be less color-sensitive.\par
For the figure and ground principle evaluation, we understand that convets have a partial discriminative capacity, however, if the scene objects are quite many, this capability is greatly deteriorated. More precisely, from Fig. \ref{plot5}, we understand that when more than four shapes are apparent in the scene, the convnet cannot infer correctly and focuses on irrelevant background features. Finally, for the symmetry evaluation, there are high peaks of performance as shown in the sensitivity analysis of Fig. \ref{plot6}. The highest peak is found in $180^{0}$ which highlights the principle of rotational object symmetry. A peak in $180^{0}$ means that the evaluated object is again symmetrical showing that the evaluated principle is apparent in the convnet.

\section{Discussion}
This paper provides a framework to analyze several underlying properties of deep convolutional networks.  The evaluation model is based on the Gestalt theory, evaluating if the principles of closure, proximity, continuation, figure and ground and symmetry are identified or violated. Preliminary results have identified several similarities between CNN mechanisms and Gestalt-like perception patterns and potential cases of their violation. There is evidence that deep models follow most of the visual cortical perceptual mechanisms defined by the Gestalt principles at several levels. The proposed systematic approach can be used for investigating further the powerful invariants of convolutional networks and their high-dimensional learning properties.

\section*{Acknowledgment}

The current postdoctoral research was fully funded by the State Scholarships Foundation (IKY) through the ``Strengthening Post-Academic Researchers'' act from the resources of the Operational Program ``Human Resources Development, Education and Lifelong Learning'' with Priority Axes 6, 8, 9 and co-funded by the European Commission Social Fund - ESF and the Greek government.


{\small
\bibliographystyle{ieee}
\bibliography{egpaper_final}
}

\end{document}